\documentclass[runningheads]{llncs}
\usepackage[T1]{fontenc}
\usepackage{graphicx}
\usepackage{hyperref}
\usepackage{color}
\renewcommand\UrlFont{\color{blue}\rmfamily}
\usepackage[inline]{enumitem}
\usepackage[noend]{algpseudocode}
\usepackage[super]{nth}
\usepackage[table]{xcolor}
\usepackage{algorithm}
\usepackage{amsfonts}
\usepackage{amsmath}
\usepackage{amssymb}
\usepackage{booktabs}
\usepackage{csquotes}
\usepackage{mathtools}
\usepackage{multirow}
\usepackage{pgfplots}
\usepackage{tikz}
\usepackage{siunitx}
\usepackage{diagbox}
\usepackage[caption=false]{subfig}
\usepackage{svg}
\usepackage{tabularx}
\usepackage{xcolor}
\usepackage{wrapfig}
\usepackage{xspace} 

\pgfplotsset{compat=1.18}
\newcommand{\falfa}{\textsc{Falfa}\xspace}

\DeclareMathOperator*{\Xtr}{\mathcal{X}_\text{train}}
\DeclareMathOperator*{\ytr}{\mathcal{Y}_\text{train}}
\DeclareMathOperator*{\Xte}{\mathcal{X}_\text{test}}
\DeclareMathOperator*{\yte}{\mathcal{Y}_\text{test}}
\DeclareMathOperator*{\Xpo}{\mathcal{X}^\prime_\text{train}}
\DeclareMathOperator*{\ypo}{\mathcal{Y}^\prime_\text{train}}
\DeclareMathOperator*{\Dpo}{\mathcal{D}^\prime_\text{train}}
\DeclareMathOperator*{\Dtr}{\mathcal{D}_\text{train}}
\DeclareMathOperator*{\Dte}{\mathcal{D}_\text{test}}

\fussy
\begin{document}

\title{Fast Adversarial Label-Flipping Attack on Tabular Data}
\titlerunning{Fast Adversarial Label-Flipping Attack}

\author{
    Xinglong Chang\inst{1} \and
    Gillian Dobbie\inst{1} \and
    J\"org Wicker\inst{1}
}

\authorrunning{X. Chang et al.}

\institute{
    University of Auckland, New Zealand \\
    \email{xcha011@aucklanduni.ac.nz, \{g.dobbie, j.wicker\}@auckland.ac.nz}
}

\maketitle

\begin{abstract}
Machine learning models are increasingly used in fields that require high reliability such as cybersecurity. However, these models remain vulnerable to various attacks, among which the adversarial label-flipping attack poses significant threats. 
In label-flipping attacks, the adversary maliciously flips a portion of training labels to compromise the machine learning model. 
This paper raises significant concerns as these attacks can camouflage a highly skewed dataset as an easily solvable classification problem, often misleading machine learning practitioners into lower defenses and miscalculations of potential risks. This concern amplifies in tabular data settings, where identifying true labels requires expertise, allowing malicious label-flipping attacks to easily slip under the radar. 
To demonstrate this risk is inherited in the adversary's objective, we propose \falfa (Fast Adversarial Label-Flipping Attack), a novel efficient attack for crafting adversarial labels. \falfa is based on transforming the adversary's objective and employs linear programming to reduce computational complexity. 
Using ten real-world tabular datasets, we demonstrate \falfa's superior attack potential, highlighting the need for robust defenses against such threats.

\end{abstract}

\section{Introduction}

Machine learning (ML) has seen extensive use in cybersecurity over recent years, particularly in intrusion detection systems, vulnerability detection, and malware classification. Despite its promising applications, ML-based security systems remain vulnerable to various adversarial attacks. Among them, data poisoning attacks present a significant threat, in which adversaries intentionally manipulate the training data, thereby leading to deteriorate the performance of ML models. In the landscape of cybersecurity-related domain, \emph{adversarial label-flipping attacks} (ALFAs) stand out. 

An ALFA is a particular type of poisoning attack where an adversary maliciously flips a portion of the training labels - this makes them the most accessible to attackers, as they merely involve tampering with training labels without having to modify the feature values. Previous work by Biggio et al.\cite{biggio2011support} and Xiao et al.\cite{xiao2012adversarial} explored the implications of ALFA on models such as Support Vector Machines (SVMs) and demonstrated the challenge of optimizing against ALFA. Later studies by Paudice et al. \cite{paudice2018label} and Zhao et al. \cite{zhang2020adversarial} also investigated ALFA on neural network models and Graph Neural Networks (GNNs), respectively.

Despite the significant literature, there exist two main gaps. Firstly, most previous work focused on image and textual data, with limited attention given to high-dimensional, mixed-type tabular data \cite{li2021detection}. Secondly, the efficiency of the ALFA algorithms has often been overlooked in the research community \cite{taheri2020defending,aryal2022analysis}. Addressing these, this paper extends prior work by introducing a fast and efficient adversarial label-flipping attack for tabular data, \falfa (Fast Adversarial Label-Flipping Attack), which proves to be a more resourceful and effective threat. 

We summarize our contributions as follows:
\begin{enumerate}
    \item The paper identifies the potential threat posed by label-flipping attacks on tabular data. It highlights how adversaries can manipulate highly skewed datasets to appear as easily solvable classification problems. This discovery is significant for security-related domains where ML practitioners may underestimate the risks associated with their training data.
    \item We propose a novel label-flipping attack called \falfa. \falfa is derived directly from the adversary's objective function. It utilizes a variable transformation technique to convert highly non-linear objective functions into linear programming problems. This transformation significantly reduces the computational time required to find the optimal subset of labels for flipping. This makes \falfa particularly effective for classifiers that optimize using the Cross-Entropy function.
    \item The paper conducts an empirical evaluation of the proposed attack on ten real-world tabular datasets. This evaluation assesses the effectiveness of \falfa on datasets of varying difficulties, providing practical insights into the performance and impact of the attack in real-world scenarios.
\end{enumerate}

\section{Methodology}
\label{sec:methodology}

\subsection{Attack Objective}
\label{sec:method:objective}

\emph{Label-flipping attacks} are a type of data poisoning attacks. 
They occur when adversaries take control of the training data.
By manipulating labels in the training data, the adversaries aim at reduce the model's performance at inference time.

We formulate the label poisoning problem as follows:
In a \emph{label-flipping attack}, the adversary can access the training data $\Dtr:=\{(\Xtr, \ytr)\}$ and has knowledge about the architecture of the model $f$ which will train on $\Dtr$ \cite{cina2022wild}. 
Based on this knowledge, the adversary replaces $\Dtr$ with a poisoned training set $\Dpo:=\{(\Xpo, \ypo)\}$.
Training on $\Dpo$ will result a poisoned model $f'$.

In a \emph{label-flipping attack}, the adversary can only alter $\ytr$, resulting $\Dpo:=\{(\Xtr, \ypo)\}$.
To increase the test error, the adversary's goal is to maximize the loss on the clean test set $\Dte:=\{(\Xte, \yte)\}$.
The objective of a \emph{label poisoning attack} \cite{munoz2017towards} is:
\begin{equation}
    \min_{\Dpo} \ell(\Dpo, f') - \ell(\Dte, f').
    \label{eq1}
\end{equation}

The optimal poisoned training set is difficult to obtain because:
\begin{enumerate}
\item \textbf{The adversary has no control on how the model is trained.} 
$\Dpo$ is designed to minimize the loss of $f'$ in order to conceal the attack from the user.
Since the model is trained to minimize $\Dpo$, it leads to the poisoned model $f'$ overfitting $\Dpo$ more easily than $\Dtr$.
This happens because the adversary can neither directly tune the parameters of $f'$ nor interfere with the normal training process.

\item \textbf{The adversary aims at maximize the classifier's error at inference time, but has no control over the data at inference time.}
The second term in Equation~\ref{eq1} is loss that the adversary wants to maximize, but the adversary does not have direct access to $\Dte$.
In order to maximize $\ell(\Dte, f')$, the adversary assumes $\Dtr$ and $\Dte$ share the same distribution, so by maximizing the loss on the original classifier $\ell(\Dpo, f)$, it indirectly affects $\ell(\Dte, f')$.
\end{enumerate}

\begin{figure}[t!]
    \centering
    \includegraphics[width=0.8\columnwidth]{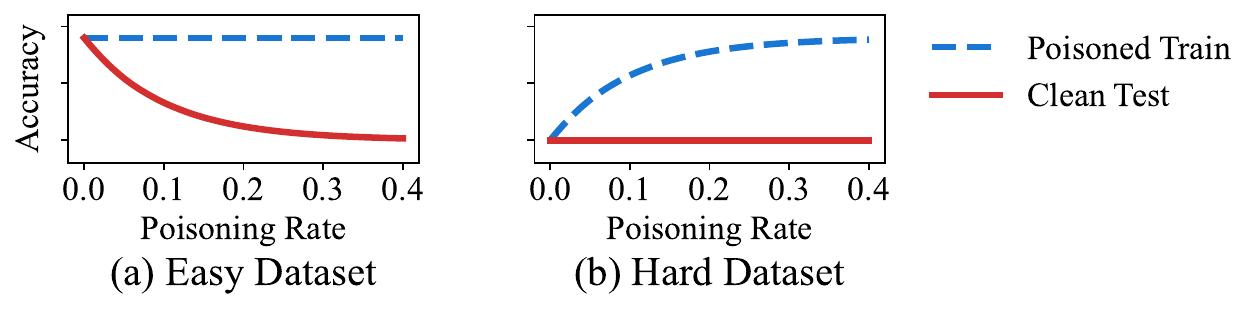}
    \caption[Theoretical Performance Degradation Under Poisoning Attacks.]{The theoretical performance degradation from poisoning attacks at various rates for easy and hard datasets.}
    \label{fig.fake}
\end{figure}

From a defender's perspective,
Equation~\ref{eq1} can be interpreted as the adversary making the discrepancy between $f$ and $f'$ as large as possible.
Since the training process optimizes $\ell(\Dpo, f')$ on poisoned data,
we know $\ell(\Dpo, f')$ must be significantly smaller than $\ell(\Dpo, f)$.
Because loss is directly linked to the model's accuracy, poisoning attacks can be detected by comparing the clean and the poisoned accuracy.

\subsection{Fast Adversarial Label-Flipping Attack}
\label{sec:method:falfa}

In this section, we introduce a novel label poisoning attack, \falfa (\emph{Fast Adversarial Label Flipping Attack}).
We directly align \falfa with the adversary's goal, making it an extremely effective untargeted attack.
This attack demonstrates that even with the lowest capability (only poisoning labels), it can have a significant negative impact on the performance of neural networks (NNs).

\falfa is derived from Equation~\ref{eq1}, the general form of an untargeted poisoning attack.
Label poisoning attacks are a special type of untargeted poisoning attack, where the adversary's capability is limited to only manipulate labels.
In a label poisoning attack, the attacker's capability is restricted to modifying labels.
In a binary classification task, given $\ytr:=\{y_i\}^n_{i=1}$ and the percentage $\epsilon$ of examples  the adversary can modify,
we hope to find the optimal solution of $\ypo:=\{y'_i\}^n_{i=1}$ to maximize the difference between $f$ and $f'$.
We rewrite Equation~\ref{eq1} with the adversary's cost constants as:
\begin{equation}
    \begin{aligned}
        \min_{\ypo} \quad   & \ell(f'(\Xtr), \ypo) - \ell(f(\Xtr), \ypo)           \\
        \textrm{s.t.} \quad & \sum_{i=1}^{n} |y'_i - y_i| \leq n \epsilon,         \\
        \quad               & y'_i \in \{0, 1\}  \; \text{for } i = 1,  \ldots, n.
    \end{aligned}
    \label{eq2}
\end{equation}
Let us consider a NN classifier, the outputs are normalized by the Softmax function:
\begin{equation}
    p_i = \text{Softmax}(x_i) = \frac{\exp{x_i}}{\sum_j\exp{x_j}}
\end{equation}
and it is optimizing the \emph{Cross Entropy Loss}:
\begin{equation}
    \ell(p_i, y_i) = y_i[-\log(p_i) + \log(1-p_i)]
\end{equation}
By expanding the objective function in Equation~\ref{eq2}, we have the following:
\begin{equation*}
    \begin{aligned}
          & \ell(f'(\Xtr), \ypo) - \ell(f(\Xtr), \ypo)                  \\
        = & \ypo[(-\log(p') + \log(1-p'))] - \ypo[-\log(p) + \log(1-p)] \\
        = & \ypo[(-\log(p') + \log(1-p')) - (-\log(p) + \log(1-p))]     \\
        = & (\alpha - \beta)\ypo
    \end{aligned}
    \label{eq3}
\end{equation*}

\noindent with $\alpha := -\log(p') + \log(1-p')$ and $\beta := -\log(p) + \log(1-p)$.
Then, Equation~\ref{eq2} becomes a linear programming problem with non-linear constraints.
Additionally, the $\beta$ term only depends on the original classifier $f$ and clean data $\Dtr$, so it is a constant and can be computed beforehand.

If Equation~\ref{eq2} were linear, it would be simple to solve, but its inequality constraint contains the absolute operation rendering it nonlinear.
We can remove this operator by considering all permutations \cite{paudice2018label},
but this approach is computationally expensive.
However, since the problem is a binary classification, we can simplify it using a multiplier
$\lambda$ with $\lambda = 1 \text{ if } y_i = 0$ and $-1$ otherwise.
Thus, $|y'_i-y_i|$ and $\lambda \cdot (y_i' - y_i)$ are equivalent, because if $y'_i=y_i$,
$\lambda$ is irrelevant.
Since $\lambda$ only depends on $\ytr$, it is a constant vector.
This transformation significantly reduces the computation cost.

We can further simplify Equation~\ref{eq2} by relaxing the boundary condition.
If we use the {\em Simplex} method to solve this linear programming problem, we can replace $y'_i \in \{0, 1\}$ by $0 \leq y'_i \leq 1$, because the solution is guaranteed on the edges.
Therefore, Equation~\ref{eq2} becomes a linear programming problem:
\begin{equation}
    \begin{aligned}
        \min_{\ypo} \quad   & (\alpha - \beta)\ypo                                      \\
        \textrm{s.t.} \quad & \lambda \cdot \ypo \leq  n \epsilon + \lambda \cdot \ytr, \\
        \quad               & 0 \leq y'_i \leq 1  \; \text{for } i = 1,  \ldots, n.
    \end{aligned}
    \label{eq4}
\end{equation}

\begin{algorithm}[t!]
    \footnotesize
    \caption{Fast Adversarial Label Flipping Attack (\falfa)}
    \begin{algorithmic}[1]
        \Require
        Original training set $\Dtr:=\{(\Xtr, \ytr)\}$, budget parameter $\epsilon$, classifier $f$ trained on $\Dtr$
        \Ensure
        Poisoned training labels $\ypo$
        \State $p \gets \text{Softmax}(f(\Xtr))$;
        \State $\beta \gets -\log(p) + \log(1-p)$;
        \State $\lambda \gets 1 \text{ if } (y_i == 1) \text{ else } -1,  \forall y_i \in \ytr$;
        \State $\ypo \gets \text{Randomly flip } n \cdot \epsilon \text{ examples on } \ytr$;
        \State $f' \gets f$;
        \While{$\ypo$ does not converge}
        \State Retrain classifier $f'$ using $(\Xtr, \ypo)$;
        \State $p' \gets \text{Softmax}(f'(\Xtr))$;
        \State $\alpha \gets - \log(p') + \log(1-p')$;
        \State Update $\ypo$ by solving Eq.~\ref{eq4};
        \EndWhile
        \State \textbf{return} $\ypo$;
    \end{algorithmic}
    \label{alg.flfa}
\end{algorithm}

The full algorithm of \falfa is shown in Algorithm~\ref{alg.flfa}.
\begin{enumerate}
    \item It is efficient on NN models. Apart from the poisoned classifier $f'$ itself, only $p'$ and $\alpha$ require updates. Anything else is constant and can be computed beforehand.
    \item It is invariant to poisoning rates. Equation~\ref{eq4} can be solved by a linear programming solver. Thus, we update $\ypo$ without looping through all permutations.
    \item It is guaranteed solvable at any poisoning rate. We randomly flip $n \cdot \epsilon$ examples on $\ytr$ and use it as the {\em initial solution} for $\ypo$. Since the initial solution is always feasible, Equation~\ref{eq4} is guaranteed to have a feasible solution.
\end{enumerate}
Noting that \falfa is not limited to NNs, it applies to any classifier that uses a \emph{Cross Entropy Loss} function.

\section{Experiments}
\label{sec:exp}

\subsection{Experimental Setup}
We repeat the experiments five times to ensure robustness.
All experiments are conducted on a workstation with 
an AMD Ryzen 9 5900 CPU with 64GB RAM and a Nvidia GeForce RTX 4090 24GB GPU running on Ubuntu 20.04.3 LTS.
The virtual environment is using \texttt{Python 3.9.18}, \texttt{PyTorch 2.0.1}, and \texttt{scikit-learn 1.3.1}.
To ensure reproducibility, all data, classifiers, hyperparameters, and code are available at \href{https://github.com/changx03/falfa23}{\UrlFont{https://github.com/changx03/falfa23}}.

\subsubsection{Datasets.\xspace}
We evaluate \falfa on 10 real-world tabular datasets from the UCI ML repository \cite{Dua2019}.
We apply normalization on all datasets during the preprocessing.
All datasets use an 80-20 training and test split.

For datasets that are multi-class classification tasks, we convert the dataset into binary based on the following:
\begin{itemize}
    \item \textbf{Abalone:} If the `Rings' attribute is less than 10, we assign the example to the negative class,  else we assign to the positive class. We exclude the categorical attribute, `Sex', and the output label, `Rings', from the inputs.

    \item \textbf{CMC:} has 3 output classes:
          \begin{enumerate*}
              \item No-use,
              \item Long-term, and
              \item Short-term
          \end{enumerate*}.
          If the class is `No-use', we assign it to the negative class, else we assign it to the positive class.

    \item \textbf{Texture:} has 10 output classes. We use a subset which contains examples labeled as `3' and `9'.
          If the class is `3', assign it to the negative class, else assign it to the positive class.

    \item \textbf{Yeast:} has 10 output classes. We select `0' and `7', the top two classes sorted by sample size.
          If the class is `0', assign it to the negative class, else assign to the positive class.
\end{itemize}

\subsubsection{Baseline Attacks.\xspace}
For ALFA and PoisSVM, we use SVM with an RBF kernel and the parameters $C$ and $\gamma$ tuned by a 5-Fold CV.
We also include {\em Stochastic Label Noise} (SLN) as a baseline attack, which randomly flips a percentage of labels.
Moreover, PoisSVM is an {\em insertion attack}, so the poisoning rate is the percentage of additional examples added to the dataset.

\subsubsection{Classifiers' Performance.\xspace}
We include \falfa, ALFA \cite{xiao2012adversarial} and Poisoning SVM (PoisSVM) \cite{biggio2012poisoning}.
For \falfa, we train a NN model with 2 hidden layers (128 neurons each) using \emph{Stochastic Gradient Descent} (SGD) for a maximum of 400 epochs, learning rate $0.01$, and mini-batch size 128.
The classifier for PoisSVM is trained using the texttt{SVC} method from the \texttt{scikit-learn} package with \emph{Radial Basis Function} (RBF) kernel with default hyper-parameters.
The baseline performance of these classifiers trained on clean data are shown in Table~\ref{tab.datasets}.

\begin{table}[ht!]
    \footnotesize
    \centering
    \caption[Summary of Real-World Data]{Summary of the training set size ($n$), number of features ($m$), Positive Label Rate (PLR), average accuracy (\%) for training and test sets across all classifiers, and difficulty (\underline{E}asy/\underline{N}ormal/\underline{H}ard).
    }
    \begin{tabular}{@{\hskip2pt}l@{\hskip2pt}|@{\hskip2pt}r@{\hskip2pt}|@{\hskip2pt}r@{\hskip2pt}|@{\hskip2pt}r@{\hskip2pt}|@{\hskip2pt}r@{\hskip2pt}|@{\hskip2pt}r@{\hskip2pt}|@{\hskip1pt}c@{\hskip1pt}}
        \toprule
        Dataset      & $n$  & $m$ & PLR  & Train Acc.    & Test Acc.     & Diff. \\
        \midrule
        Abalone      & 1600 & 7   & 0.50 & $79.9\pm0.7$  & $76.5\pm0.5$  & N     \\
        Australian   & 552  & 14  & 0.45 & $91.5\pm3.1$  & $81.9\pm2.1$  & N     \\
        Banknote     & 1097 & 4   & 0.44 & $100.0\pm0.0$ & $100.0\pm0.0$ & E     \\
        Breastcancer & 455  & 30  & 0.63 & $99.3\pm0.2$  & $95.0\pm2.5$  & E     \\
        CMC          & 1178 & 9   & 0.77 & $79.9\pm2.8$  & $77.5\pm0.6$  & N     \\
        HTRU2        & 1600 & 8   & 0.50 & $94.8\pm0.5$  & $92.6\pm0.9$  & E     \\
        Phoneme      & 1600 & 5   & 0.50 & $89.7\pm6.3$  & $85.6\pm1.3$  & N     \\
        Ringnorm     & 1600 & 20  & 0.50 & $99.4\pm0.4$  & $97.8\pm1.1$  & E     \\
        Texture      & 800  & 40  & 0.50 & $100.0\pm0.0$ & $99.8\pm0.5$  & E     \\
        Yeast        & 713  & 8   & 0.48 & $73.5\pm4.7$  & $65.8\pm1.6$  & H     \\
        \bottomrule
    \end{tabular}
    \label{tab.datasets}
\end{table}

\subsection{Experimental Results}
We present the performance loss at a poisoning rate of $10\%$ in Table~\ref{tab.err10}.
This table shows test accuracy differences before and after the attack.
At the poisoning rate of $10\%$, SLN has no meaningful impact on NN models' performance with an average performance loss of $0.98\%$,  demonstrating that NNs are robust against SLN at $10\%$.
\falfa and ALFA are tied in top-ranked attacks with 7 out of 10 datasets on paired two sample T-tests with $\alpha$ set to $0.1$.
\falfa outperforms ALFA by a large margin on the Phoneme and Ringnorm datasets but is bested by ALFA on the Abalone and Yeast datasets.

\begin{table}[ht!]
    \footnotesize
    \centering
    \caption[Performance Loss Under Label Poisoning Attacks]{Performance loss (\%) after attacked by a poisoning attack with 10\% poisoning rate. Top ranked attacks are marked in \textbf{bold}.}
    \begin{tabular}{l|@{\hskip2pt}r@{\hskip2pt}|@{\hskip2pt}r@{\hskip2pt}|@{\hskip2pt}r@{\hskip2pt}|@{\hskip2pt}r@{\hskip2pt}}
        \toprule
        Dataset      & SLN          & PoisSVM               & ALFA                  & \falfa                \\
        \midrule
        Abalone      & $0.8\pm0.7$  & $1.8\pm0.8$           & $\mathbf{9.5\pm1.9}$  & $7.7\pm1.7$           \\
        Australian   & $0.7\pm0.5$  & $4.5\pm3.9$           & $\mathbf{4.9\pm4.0}$  & $\mathbf{8.3\pm3.8}$  \\
        Banknote     & $1.4\pm2.3$  & $1.1\pm1.1$           & $\mathbf{10.9\pm2.5}$ & $\mathbf{10.3\pm2.9}$ \\
        Breastcancer & $2.5\pm0.7$  & $5.3\pm4.6$           & $\mathbf{7.2\pm2.0}$  & $\mathbf{9.1\pm2.7}$  \\
        CMC          & $-0.2\pm0.7$ & $\mathbf{15.1\pm4.7}$ & $3.5\pm3.0$           & $5.7\pm3.3$           \\
        HTRU2        & $0.7\pm0.3$  & $0.7\pm1.3$           & $\mathbf{9.2\pm3.1}$  & $\mathbf{9.4\pm2.4}$  \\
        Phoneme      & $3.5\pm2.9$  & $0.9\pm2.1$           & $6.8\pm0.7$           & $\mathbf{11.6\pm2.1}$ \\
        Ringnorm     & $0.1\pm0.3$  & $1.7\pm0.5$           & $3.2\pm2.5$           & $\mathbf{6.4\pm2.9}$  \\
        Texture      & $0.5\pm1.1$  & $1.2\pm0.8$           & $\mathbf{7.9\pm4.6}$  & $\mathbf{4.9\pm3.9}$  \\
        Yeast        & $-0.2\pm1.6$ & $1.9\pm3.8$           & $\mathbf{10.4\pm4.9}$ & $2.3\pm4.6$           \\
        \bottomrule
    \end{tabular}
    \label{tab.err10}
\end{table}

\begin{figure}[ht!]
    \centering
    \includegraphics[width=\columnwidth]{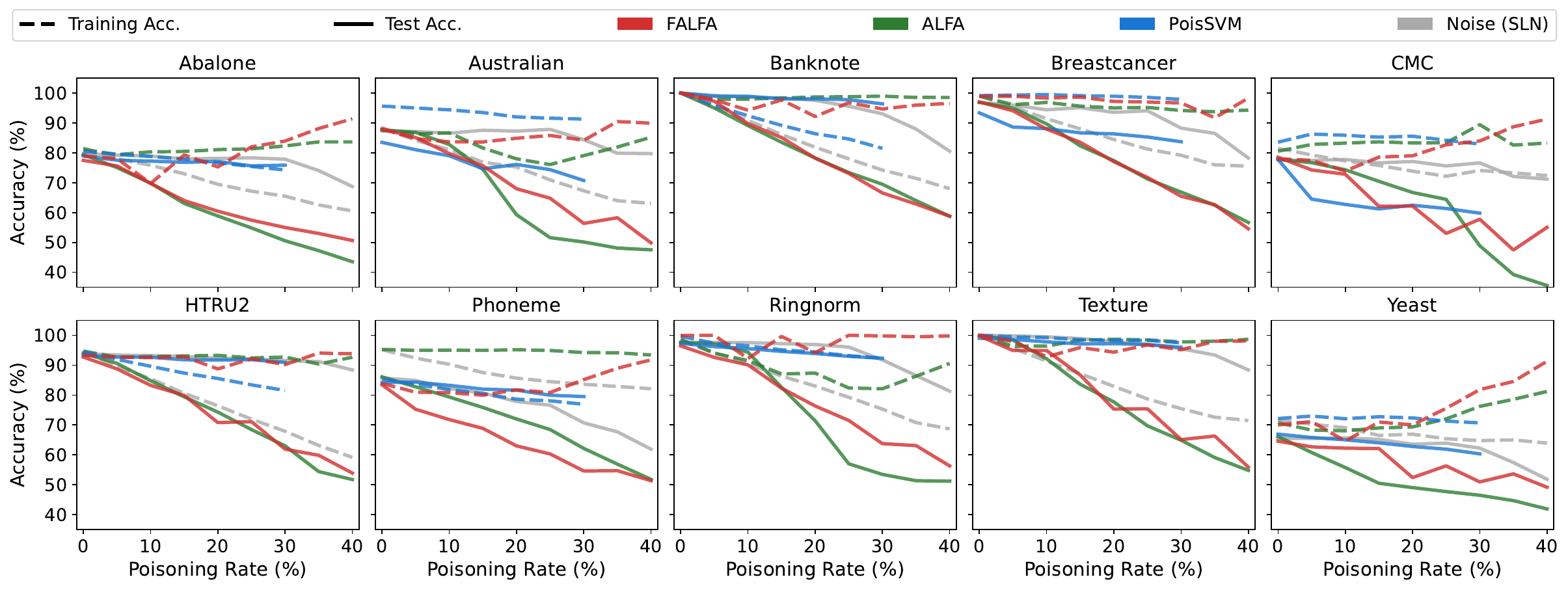}
    \caption[Training and Test Accuracy Under Poisoning Attacks]{
        The training and test accuracy at various poisoning rates exhibit similar patterns under different attacks on the same dataset. However, the difficulty of the classification task has a large influence on the behavior of attacks.
    }
    \label{fig.real_acc}
\end{figure}

Figure \ref{fig.real_acc} illustrates the relationship between the poisoned training accuracy and the clean test accuracy when a classifier is poisoned at various rates\footnote{We are only able to run PoisSVM up to 30\% poisoning rate using the implementation provided by the original paper \cite{biggio2012poisoning}.}.
We observe that \falfa and ALFA reliably degrade the classifier performance on all three datasets. However, PoisSVM is substantially weaker.

Previous works use the test classification error as the sole performance metric to benchmark the poisoning attacks \cite{ho2002complexity,koh2022stronger,paudice2018label}.
However, this does not fully reflect the impact of poisoned labels.
To conceal an attack from the user, keeping the training accuracy (dashed lines in Figure \ref{fig.real_acc}) stationary is equally essential.
The behavior of SLN (gray lines in Fig.~\ref{fig.real_acc}) is different from other attacks, as the training and test accuracy fall at a similar rate, highlighting the key motivation of this research.

Given a difficult classification task,
an increasing poisoning rate has a limited impact on the testing accuracy, such as the Yeast dataset in Fig.~\ref{fig.real_acc}.
Meanwhile, the training accuracy on poisoned data rises.
This pattern is observed in poisoning attacks, but not in noise.
We believe this is due to the poisoning attacks optimizing the second term in Equation~\ref{eq2}.
When the testing accuracy is close to random levels, an attack algorithm can no longer maximize the loss on test data; it will minimize the loss on the training data instead.
Thus, the classifier becomes more likely to overfit the poisoned training set.

\subsubsection{Time Complexity.\xspace}

\falfa is more computationally efficient than ALFA and PoisSVM by a substantial margin.
Linear programming is an exponential-time algorithm, where the time complexity is around $O(n^{2.5})$.
Xiao {\em et al.}'s ALFA creates a copy of $\ytr$ in the linear programming step, so $n$ is essentially doubled.
Paudice {\em et al.}'s ALFA on NNs is slower than Xiao {\em et al.}'s, since it traverses all combinations of $\ytr$ instead of using linear programming.
\falfa uses linear programming but without doubling $\ytr$, resulting in an approximately $2^{2.5}\approx5.6$ times faster than ALFA on each iteration.
Our test shows that \falfa converges at 2 iterations on average, but ALFA takes more than 5 iterations to converge.
In the worst-case scenario, \falfa poisons the CMC dataset in $22.4\pm8.6$ secs, while ALFA requires $405.8\pm348.4$ secs, and PoisSVM took over 2 hours.
We observe the minimal difference on BreastCancer, where \falfa completes the task at $5.3\pm1.9$ secs, and it takes ALFA $7.4\pm5.6$ secs.

\section{Conclusion and Future Work}
\label{sec:conclusion}

In this paper, we have demonstrated that by applying label-flipping attacks, adversaries can camouflage a highly skewed dataset as an easily solvable classification problem. 
This poses a significant concern in security-related domains, where ML practitioners may inadvertently lower their guard and miscalculate the potential risks associated with the training data.
To illustrate that this behavior aligns with the adversary's objectives, we introduce \falfa.
\falfa is directly derived from the adversary's objective function. 
Leveraging the variable transformation technique, \falfa converts the highly non-linear objective function into a linear programming problem. 
This transformation significantly reduces the computational time required to search the optimal subset of labels for flipping, making it particularly effective for classifiers that optimize using the Cross-Entropy function.
In our future work, we plan to develop defenses capable of identifying such attacks, even when poisoned labels attempt to disguise as legitimate data.

\bibliographystyle{splncs04}
\bibliography{references}

\end{document}